\documentclass[10pt,twocolumn,letterpaper]{article}
\usepackage[pagenumbers]{cvpr}
\usepackage{microtype}
\usepackage{placeins}
\usepackage{flushend}
\usepackage{needspace}
\definecolor{cvprblue}{rgb}{0.21,0.49,0.74}
\usepackage[breaklinks,colorlinks,allcolors=cvprblue]{hyperref}
\title{CoRA-VTLA: Contact-Guided Attention and Queryable Response Memory for Tactile Post-Training}
\author{%
    Chengye Song\textsuperscript{1}\enspace
    Rui Song\textsuperscript{1}\enspace
    Shengqi Wang\textsuperscript{1}\enspace
    Jiawei Zhang\textsuperscript{2}\enspace
    Hongzhou Wang\textsuperscript{2,4}\enspace
    Xueyang Guo\textsuperscript{3}\enspace
    Yang Liu\textsuperscript{1}\\[4pt]
    \textsuperscript{1}Dalian University of Technology\quad
    \textsuperscript{2}Jilin University\quad
    \textsuperscript{3}Tsinghua University\quad
    \textsuperscript{4}Enomind Technology%
}

\makeatletter
\renewcommand{\@maketitle}{%
    \null
    \begingroup
    \centering
    {\LARGE\bfseries \@title\par}
    \medskip
    \begin{tabular}{@{}c@{}}\@author\end{tabular}\par
    \endgroup
    \medskip
    \noindent\textbf{Abstract.}\enspace
    In precision manipulation, small errors in alignment or force can cause jamming or deformation, making contact feedback essential for adapting pretrained vision--language--action (VLA) models to these tasks. A single tactile observation leaves the response to recent actions ambiguous. Direct tactile fusion during post-training can also disperse the action expert's visual attention and reduce the share it allocates to contact regions. To address these challenges, CoRA-VTLA augments pretrained VLAs with lightweight tactile modules. A queryable action--response memory retrieves recent contact responses so that the action expert can combine them with immediate tactile feedback. Contact-guided gaze supervision concentrates the visual attention of the tactile semantic token on image regions corresponding to fingertip contact. CoRA-VTLA achieves 73.33\% mean success across three MuJoCo tasks, compared with 50\% for direct tactile fusion with the same sensor inputs. It also achieves 67\% success on an optical-tactile task from the public UniVTAC benchmark and 66.7\% mean success across three real-robot tasks. These results show that tactile post-training helps pretrained VLAs adjust to changing contact.
    \par\smallskip
    \noindent\textbf{Project page.} \url{https://cora-vtla.github.io/}\par
    \medskip
    \noindent\begin{minipage}{\textwidth}
        \centering
        \captionsetup{type=figure}
        \includegraphics[width=\textwidth]{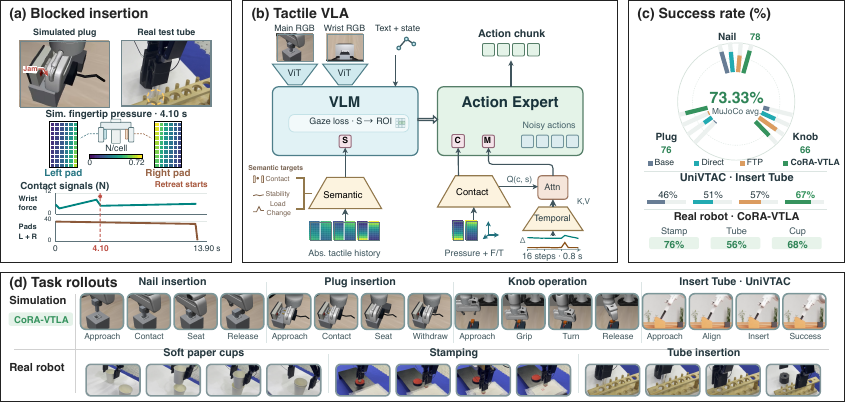}
        \caption{CoRA-VTLA at a glance. \textbf{(a)} Blocked insertions in simulation and on the real robot, with simulated fingertip pressure maps and load traces around withdrawal. \textbf{(b)} A tactile semantic token enters the VLM, while current contact and queried action--response memory condition the action expert. \textbf{(c)} Success rates in MuJoCo and on UniVTAC Insert Tube, along with CoRA-VTLA's real-robot success rates. \textbf{(d)} Keyframes from simulated and real-robot rollouts.}
        \label{fig:teaser}
    \end{minipage}\par
    \vspace{8pt}
}
\makeatother

\begin{document}
\maketitle

\section{Introduction}
\label{sec:introduction}

Pretrained vision-language-action (VLA) models reuse visual and linguistic knowledge across manipulation tasks~\cite{rt2,openvla,pi05}. Yet precision manipulation requires corrections after contact: small alignment errors can jam assemblies, while excessive force can deform parts or fixtures~\cite{lift,univtac}. Visual observations guide approach and alignment but cannot reliably distinguish free motion from hidden resistance once parts touch. This makes post-contact adjustment crucial in assembly and delicate handling. In peg insertion, combining visual feedback with wrist force/torque measurements improves policy learning~\cite{makingSense}. In Fig.~\ref{fig:teaser}(a), a part appears aligned but stops advancing against a fixture. Tactile sensing reveals such interactions and supports corrective action. Tactile data are scarce in VLA pretraining. Task demonstrations with tactile observations offer a practical route to post-training for contact-rich manipulation~\cite{simpletouch}.\par
\Needspace{2\baselineskip}
Force-aware control uses measured loads to guide motion~\cite{forcevla2}. Tactile prediction provides supervision for learning contact dynamics~\cite{simpletouch,tacwam}. Force histories have also been used to track task progress through repeated contact events~\cite{fmvla}. The challenge during recovery is to determine which recent responses matter for the next correction. For example, initial engagement and a sustained jam can produce similar instantaneous loads, yet call for continued insertion or withdrawal and realignment (Fig.~\ref{fig:plug-recovery}). Prior work uses executed motion to interpret sensor measurements~\cite{interactiveperception} and predict tactile signals during slip~\cite{actiontactileprediction}. For recovery, we consider how past contact responded to motion and which responses matter in the current state. How can a pretrained policy retrieve the contact responses most relevant to its next action?

Tactile post-training also challenges visual grounding. Prior work uses the location of touch to relate local readings to the full visual scene~\cite{crossmodalprediction} and uses tactile geometry to localize grasped parts for insertion~\cite{gelsightinsertion}. Direct tactile injection can disrupt visual dynamics modeling in world action models~\cite{tactilewam}. In VLAs, direct tactile fusion can draw action-expert attention away from the target, motivating adaptive tactile injection~\cite{atvla}. On matched MuJoCo observations, direct tactile fusion likewise produces more dispersed main-camera attention for the first action token at layer 17 (Sec.~\ref{sec:qualitative}). Regulating tactile injection leaves a spatial question open: which image regions correspond to the sensed contact? Reliable correction therefore requires both identifying where contact occurs and retrieving how it has evolved in response to motion.

To address these challenges, \textbf{CoRA-VTLA} is proposed as a tactile post-training framework combining response-aware action generation with spatially grounded tactile perception. At the action level, a queryable action--response memory encodes measured contact changes and retrieves responses conditioned on current contact and robot state. At the perception level, contact-guided gaze supervision directs a tactile semantic token's visual attention toward projected fingertip contact regions. The two mechanisms support different aspects of contact-aware correction within one pretrained VLA. The action expert combines retrieved responses, current feedback, and VLM task context to generate the next action chunk. The main contributions are summarized as follows:
\begin{itemize}
    \item \textbf{Queryable action--response memory.}
    CoRA-VTLA introduces state-conditioned retrieval of measured contact responses for corrective action generation, with next-step response prediction on task demonstrations to initialize action-side tactile encoders.

    \item \textbf{Contact-guided gaze supervision.}
    Projected fingertip contacts provide spatial supervision for grounding tactile semantic representations in visual contact regions during VLA post-training.

    \item \textbf{Simulation and real-robot evaluation.}
    CoRA-VTLA achieves 73.33\% mean success across three custom MuJoCo tasks, compared with 50\% for direct tactile fusion with matched inputs. The memory design reaches 67\% on a public UniVTAC optical-tactile task~\cite{univtac}. The full framework averages 66.7\% across three real-robot tasks.
\end{itemize}

\section{Related Work}
\label{sec:related-work}

\subsection{Pretrained Robot Policies}

Large-scale demonstrations and cross-embodiment datasets support reusable robot policies~\cite{rt1,oxe,octo}. Embodied language models incorporate continuous sensor inputs~\cite{palme}. Vision-language-action (VLA) models transfer visual and linguistic priors to action prediction~\cite{rt2,openvla}. Action chunking, diffusion, and flow matching provide different ways to model continuous robot actions~\cite{act,diffusionpolicy,pi0,pi05}. Optimized fine-tuning improves adaptation to downstream tasks~\cite{openvlaoft}. For policies pretrained primarily on visual, linguistic, and proprioceptive inputs, using measured contact responses to correct actions in closed loop remains a challenge during task-specific adaptation.

\subsection{Tactile Representations and Policies}

Robotic tactile sensing uses interfaces ranging from optical deformation sensors to magnetic soft skins~\cite{gelsight,digit,reskin}. Paired visual and tactile data support cross-modal prediction and multimodal representation learning~\cite{crossmodalprediction,touchgo,makingSense}. Self-supervised approaches learn tactile representations from robot interaction~\cite{tdex,sparsh,sparshskin}. Work on tactile representation learning studies transfer across sensors~\cite{t3,anytouch} and alignment of touch with vision and language~\cite{unitouch,tvl}. Force supervision provides additional cues about physical interaction~\cite{tafvla}.

At the policy level, tactile feedback supports grasp adjustment, and visual rewards can guide the learning of tactile skills~\cite{calandra,tavi}. Tactile VLAs incorporate contact information through token fusion, contact gating, force-aware action experts, and adaptive feedback~\cite{forcevla,tacvla,tactilevla,atvla,vlatouch,lift}. Training approaches range from large-scale tactile policy pretraining to predictive supervision on task demonstrations~\cite{ftp1,n0vtla,simpletouch}. These approaches give policies access to contact information. During recovery, a remaining challenge is to determine whether the sensed contact warrants a correction and what that correction should be.

% Queue Fig. 2 for the page where Method begins.
\begin{figure*}[t]
    \centering
    \includegraphics[width=\textwidth]{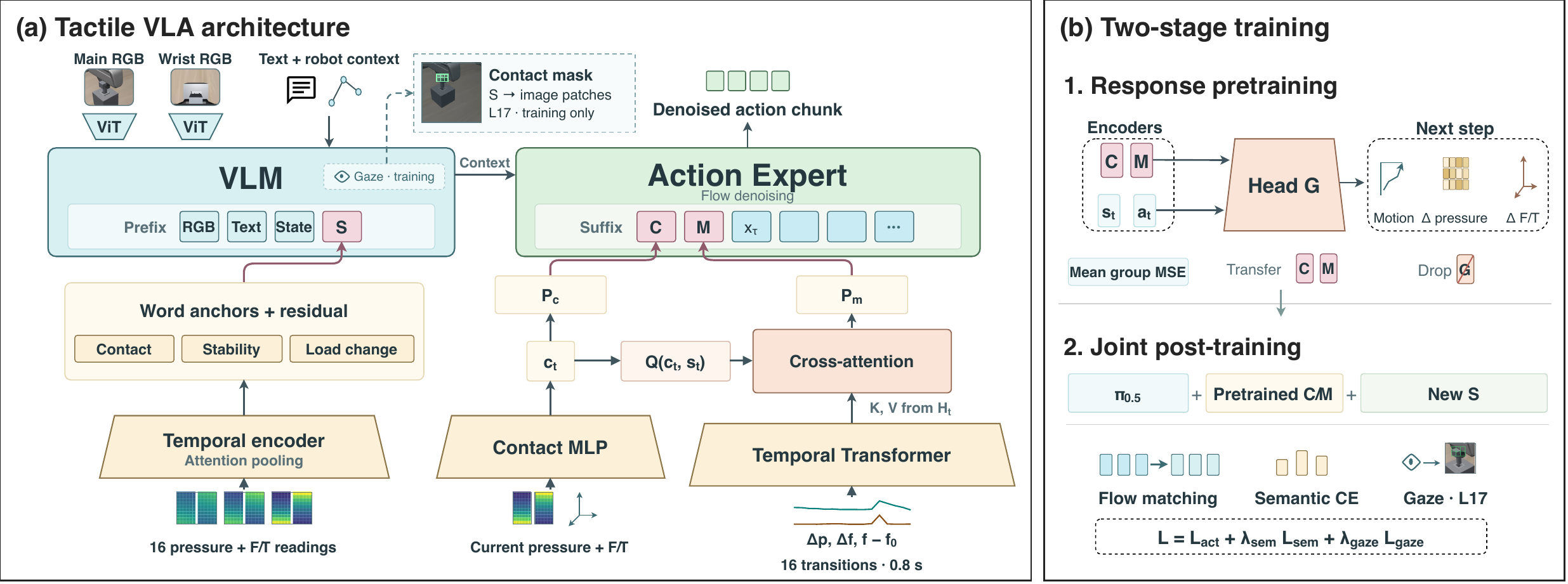}
    \caption{CoRA-VTLA architecture and training. \textbf{(a)} A semantic token $S$ joins the VLM prefix. The contact and memory tokens $C$ and $M$ join the action-expert suffix. \textbf{(b)} Response prediction initializes the contact and memory encoders, followed by joint post-training with action, semantic, and gaze supervision.}
    \label{fig:method}
\end{figure*}

\subsection{Temporal Prediction and Spatial Grounding}

Tactile dynamics models support predictive control and slip prediction~\cite{tactilempc,actiontactileprediction}. ViTacFormer combines cross-modal fusion with autoregressive tactile prediction~\cite{vitacformer}. Visuotactile diffusion policies use temporal observations and auxiliary prediction of tactile changes~\cite{auxiliarytactiledp}. Recent world--action models use tactile histories and future supervision~\cite{tacwam}. Other approaches learn predictive contact states~\cite{tacmind}. Some combine prediction with reactive feedback~\cite{touchworld,tacpac}. Selecting which recent measured contact responses should inform the next action remains a challenge for closed-loop control.

Spatial grounding links tactile feedback to visual information. Visuo-Tactile Transformers use tactile inputs to emphasize task-relevant visual features~\cite{vtt}. ReconVLA encourages visual grounding by reconstructing regions around manipulated objects~\cite{reconvla}. In precision manipulation, a further challenge is to associate a specific fingertip measurement with the image region where that contact occurs.

\section{Method}
\label{sec:method}

As shown in Fig.~\ref{fig:method}, CoRA-VTLA conditions the action expert on current contact and responses retrieved from the action--response memory, while contact-guided gaze supervision aligns a tactile semantic token with the corresponding image regions. Training first uses response prediction to initialize the contact and memory encoders, then jointly optimizes the policy with action, semantic, and gaze supervision.

\subsection{Overview}

CoRA-VTLA adapts a pretrained VLA policy to contact-rich manipulation using synchronized task demonstrations. At step $t$, the policy receives RGB images, a language instruction, robot state $s_t$, recent robot history, fingertip pressure $p_t$, and wrist force/torque $f_t$.

A pressure map captures how the fingers are loaded, while the wrist wrench measures the net loading on the gripper. Similar instantaneous readings can arise as the tool presses against an obstruction or as contact settles. Recent responses distinguish these interaction histories. As shown in Fig.~\ref{fig:method}(a), the action expert receives two tokens that encode the current load and retrieve recent responses. A third token supplies tactile semantics to the VLM. The policy predicts an action chunk, executes its first command, and replans.

\subsection{Contact \& Interaction Memory}

\noindent \textbf{Current contact.} The encoder $E_c$ combines fingertip pressure and wrist wrench into a contact feature $c_t=E_c([p_t;f_t])$. The projection $P_c$ maps this feature to the action expert's token space.
\begin{equation}
\label{eq:method2}
z_t^c=P_c\!\left(E_c([p_t;f_t])\right).
\end{equation}
The contact encoder receives the fingertip map and wrist wrench together. The map retains the distribution across both fingers, so a load concentrated on one side and a balanced load can yield different contact features even when their total pressure is similar.

\noindent \textbf{Action--response memory.} The memory records the load response measured after each executed action. Each event $r_i=[\Delta p_i;\Delta f_i;f_i-f_0]$ combines the changes in pressure and wrist wrench between consecutive observations with the wrist wrench relative to its initial reading $f_0$. The differences capture the load change after each executed command. The episode-relative wrench retains sustained loading even when the stepwise difference becomes small. At step $t$, the window ends with the response already observed at $t$. In MuJoCo, the most recent $K=16$ events span 0.8 s.

A temporal Transformer encodes the ordered events as $H_t=\mathcal T(r_{t-K+1:t})$~\cite{transformer}. The query $q_t=Q([c_t;s_t])$ combines the current contact feature and robot state. The query reads $H_t$ as keys and values. The resulting memory token is
\begin{equation}
\label{eq:method4}
z_t^m=P_m\!\left(\operatorname{Attn}(q_t,H_t,H_t)\right).
\end{equation}
During insertion, a tool still pressing against a rim and one re-establishing contact after backing off can produce similar instantaneous loads. Their recent responses differ: a drop in load during withdrawal can indicate that the tool has cleared the obstruction, and a subsequent rise can signal renewed contact. Conditioning the query on the current pose and load lets the action expert use the part of that history relevant to its next move.

\subsection{Semantic Grounding and Policy Integration}

\noindent \textbf{Action-side integration.} The contact and memory tokens precede the noisy action tokens in the $\pi_{0.5}$ action-expert suffix. The expert denoises the action chunk using both tactile tokens and the context from the VLM prefix. The two tokens let the expert read the instantaneous load and its recent evolution separately while predicting the same chunk.

\noindent \textbf{Tactile semantics.} A parallel temporal encoder pools recent absolute pressure and wrist readings into $v_t$. Three heads predict contact, stability, and load change. For each of the $M=3$ properties, the probabilities $\pi_k(v_t)$ weight the two word embeddings in $E_k$. The semantic token for the VLM prefix is the mean of these weighted embeddings plus a residual projection.
\begin{equation}
\label{eq:method5}
z_t^{\mathrm{sem}}=\frac{1}{M}\sum_{k=1}^{M}E_k\pi_k(v_t)+P_\delta(v_t),
\end{equation}
Training labels for the three heads are derived directly from tactile measurements. Total fingertip load determines whether contact is present. During sustained contact, the magnitude of recent pressure and wrist-wrench changes determines stability. A load change is flagged when fingertip pressure rises or the wrist wrench varies beyond a threshold. The word anchors express these properties in the VLM embedding space, and the residual carries information beyond the three predictions.

\noindent \textbf{Contact-region grounding.} AT-VLA observed that direct tactile fusion can shift the action expert's attention away from the target object~\cite{atvla}. The tactile semantic token assigned little attention to the sensed contact region when trained without spatial supervision (Sec.~\ref{sec:qualitative}). To guide the token toward contact, a contact region $R_t$ is defined by projecting active fingertip cells onto the main camera's patch grid, using the standard Vision Transformer image-patch representation~\cite{vit}. Let $\alpha_{t,j}$ be the semantic token's attention to patch $j$ at the penultimate VLM layer, normalized across the main camera's patches. For valid projections, the gaze loss penalizes attention mass in $R_t$ below $\eta$, where $(x)_+=\max(0,x)$.
\begin{equation}
\label{eq:method6}
\mathcal L_{\mathrm{gaze}}=\mathbb E\!\left[
\left(\eta-\sum_{j\in R_t}\alpha_{t,j}\right)_+^2\right].
\end{equation}
The semantic labels describe the state of contact, and the projected region locates that contact in the image. As the tool moves, the region follows the measured fingertip contacts. The projected region serves as a training target for the semantic token's visual attention. Once the region receives the target share, the gaze penalty is zero. The token remains free to attend to nearby geometry, such as the fixture opening. At inference, it attends to the full image through the VLM.

\subsection{Response Pretraining and Joint Learning}

\noindent \textbf{Response pretraining.} The contact and memory encoders are pretrained on the task demonstrations (Fig.~\ref{fig:method}(b)). At the same contact state, advancing against a rim can build up load, while backing off can release it. A temporary head $G$ is therefore conditioned on the demonstrated command $a_t$ alongside the two tactile tokens and the robot state $s_t$. Its prediction $\hat y_{t+1}$ has three groups $\mathcal G$: realized motion, pressure change, and wrist-wrench change. Each group is supervised by the response measured after command execution.
\begin{equation}
\label{eq:method7}
\mathcal L_{\mathrm{resp}}=\frac{1}{3}\sum_{g\in\mathcal G}
\operatorname{MSE}(\hat y_{t+1}^{g},y_{t+1}^{g}).
\end{equation}
The command conditions the response head, so the encoders learn features that help predict the consequences of different actions from observations available to the policy. Averaging the group losses gives motion, pressure, and wrench changes equal weight regardless of their channel counts. This stage adds little computational overhead because it trains only the two tactile encoders and a small prediction head on the same demonstrations, without running the VLA backbone. The two encoders are retained for joint post-training, and the prediction head is discarded.

\noindent \textbf{Joint post-training.} Joint post-training starts from pretrained $\pi_{0.5}$~\cite{pi05} and the two response-pretrained encoders, and updates the visual backbone, VLM, action expert, and tactile modules, including the newly initialized semantic branch. The objective combines a tactile property classification loss, the gaze loss, and a flow-matching action loss~\cite{flowmatching}.
\begin{equation}
\label{eq:method9}
\mathcal L_{\mathrm{joint}}=\mathcal L_{\mathrm{act}}
+\lambda_{\mathrm{sem}}(k)\mathcal L_{\mathrm{sem}}
+\lambda_{\mathrm{gaze}}(k)\mathcal L_{\mathrm{gaze}}.
\end{equation}
Here $k$ indexes the training updates.

\begin{figure*}[t]
    \centering
    \includegraphics[width=\textwidth]{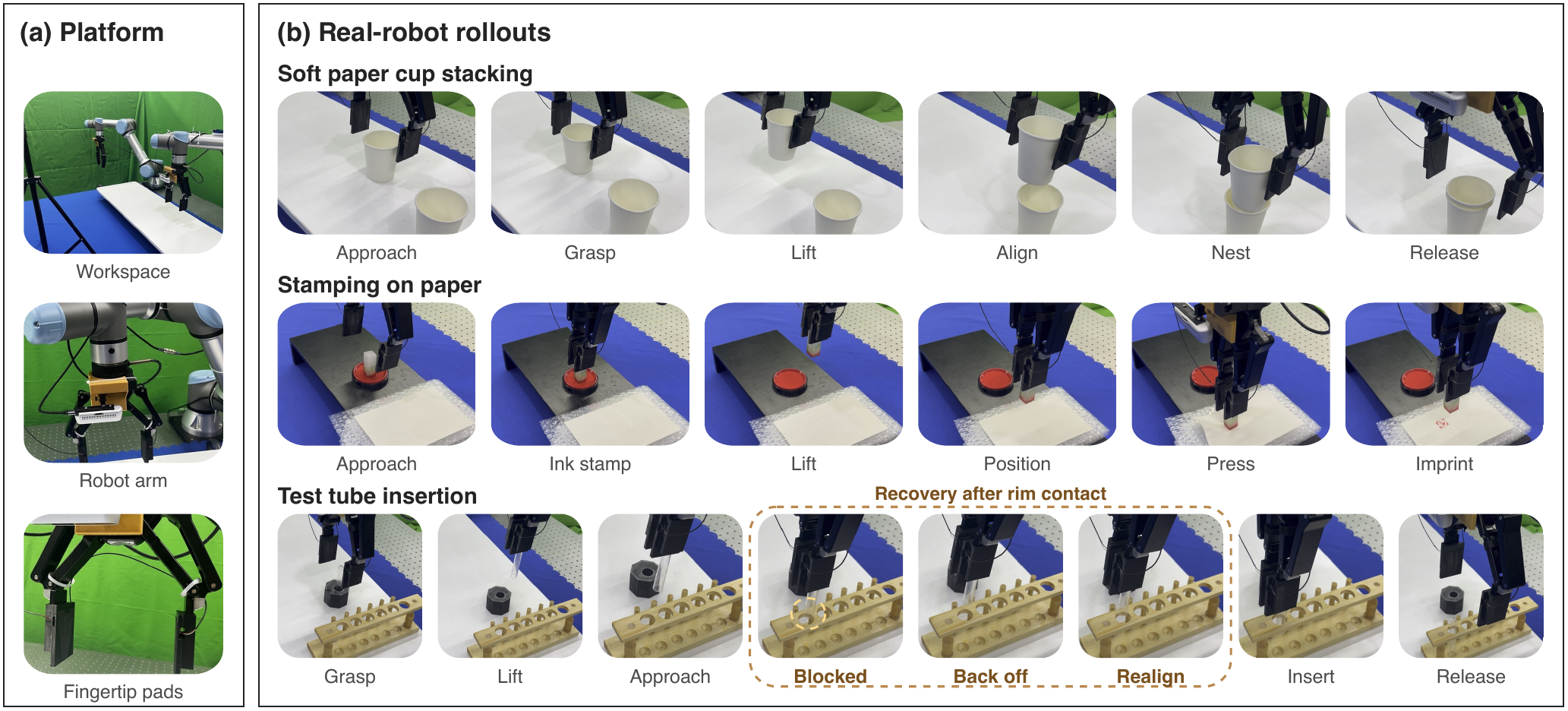}
    \caption{Real-robot platform and tasks. \textbf{(a)} Workspace, robot arm, and fingertip pads. \textbf{(b)} Keyframes of CoRA-VTLA stacking soft disposable paper cups, stamping on paper, and inserting a test tube into a rack. The highlighted frames show the tube catching on the rim before the robot backs off and realigns to complete the insertion.}
    \label{fig:real-platform}
\end{figure*}

\Needspace{5\baselineskip}
\section{Experiments}
\label{sec:experiments}

The evaluation covers three custom MuJoCo tasks, three real-robot tasks, and Insert Tube on the public UniVTAC benchmark. The main comparisons assess task success and test whether action--response memory can be adapted to optical tactile sensing. Ablation studies examine the roles of the memory and training objectives, while qualitative analyses illustrate recovery from blocked insertion and attention to the contact region.

\subsection{Experimental Setup}

\noindent \textbf{Simulation tasks and demonstrations.} Evaluation covers three contact-rich tasks in MuJoCo~\cite{mujoco}: nail insertion, plug insertion, and knob operation. The insertion tasks require the robot to grasp a small workpiece, transport it to a fixture, and complete the insertion with a stable release, while the knob task requires pressing the knob to unlock and rotating it to a target angle under varying press stiffness and grasp friction. Each task provides demonstrations of both nominal execution and recovery from blocked interactions, with scene initial conditions and physical parameters varying within each task. A single policy is jointly trained on all three tasks. Tactile sensing provides two fingertip normal-pressure arrays and six-axis wrist force/torque, as defined in Sec.~\ref{sec:method}.

\noindent \textbf{Real-robot tasks.} The platform uses a UR5e robot arm with a Robotiq 140 gripper equipped with Xuanya electronic skin on both fingertips (Fig.~\ref{fig:real-platform}). In cup stacking, the robot grasps a soft paper cup, nests it inside a second cup, and releases it stably. The grip must limit deformation, and off-center rim contact can displace the lower cup. In stamping, the robot inks the stamp and presses it onto paper to leave a complete imprint. Variations in grasp height change when the stamp touches the paper and how much downward travel is needed. In tube insertion, the robot seats a test tube in the target hole and releases it stably. If the tube catches on the rim, recovery requires backing off and realigning. Each method is evaluated in 50 trials per task. Demonstrations, initial-condition distributions, and task-specific success and safety criteria are shared across compared methods.

\noindent \textbf{Compared methods.} The comparison includes Base, Direct, adapted FTP-$\pi_{0.5}$, and CoRA-VTLA. All methods share RGB images, language instructions, robot state, and causal robot-history inputs. Base uses these common inputs alone. Direct encodes current pressure and wrist force/torque into one action-side token. FTP-$\pi_{0.5}$~\cite{ftp1} adopts the authors' tactile encoding and expert mechanism, adapted to the common sensors and action interface. CoRA-VTLA uses the complete framework in Sec.~\ref{sec:method}.

\noindent \textbf{Training protocol.} For the three MuJoCo tasks, all methods are initialized independently from the same generic $\pi_{0.5}$ checkpoint and fine-tuned with full parameters under an identical update budget and sampling schedule; the action-side tactile modules of CoRA-VTLA are first initialized by response pretraining on the same demonstrations. Peak learning rates are $2\times10^{-6}$ for vision, $5\times10^{-6}$ for the VLM, $2.5\times10^{-5}$ for the action expert, and $7.5\times10^{-5}$ for the added tactile parameters, each with warmup followed by cosine decay. Normalization statistics and auxiliary supervision are derived only from training data, with consistent temporal indexing between training and online inference.

\noindent \textbf{Closed-loop evaluation.} Each MuJoCo policy is evaluated on the same 50 unseen scenes per task, evenly split between nominal and recovery initial conditions, for 150 trials in total, using its own final checkpoint. Policies run at 20 Hz with 500 Hz physics, predicting 50-command chunks with 10 ODE steps and executing the first command before replanning, with at most 600 policy commands per episode. Insertion success requires valid insertion geometry, release, gripper withdrawal, and eight consecutive stable released observations; knob success requires the target rotation, release of the press and grasp, gripper withdrawal, and the same stability criterion. Penetration above 2 mm or gripper--fixture contact causes a safety failure, and knob press travel has an additional safety bound. Safety failures, timeouts, and initialization failures are retained in the success-rate denominator.

\subsection{Main Results}

\begin{table}[t]
\centering
\caption{Success rates (\%) on the three MuJoCo tasks and UniVTAC Insert Tube. Bold marks the best result in each column.}
\label{tab:simulation}
\small
\setlength{\tabcolsep}{3.5pt}
\begin{tabular}{lrrrrr}
\toprule
 & \multicolumn{4}{c}{MuJoCo tasks} & UniVTAC \\
\cmidrule(lr){2-5} \cmidrule(lr){6-6}
Method & Nail & Plug & Knob & Avg. & Insert Tube \\
\midrule
Base ($\pi_{0.5}$) & 74 & 0 & 20 & 31.33 & 46 \\
Direct & 68 & 32 & 50 & 50.00 & 51 \\
FTP-$\pi_{0.5}$ & 56 & 4 & \textbf{72} & 44.00 & 57 \\
\midrule
\textbf{CoRA-VTLA} & \textbf{78} & \textbf{76} & 66 & \textbf{73.33} & \textbf{67} \\
\bottomrule
\end{tabular}
\end{table}

\noindent \textbf{MuJoCo results.} CoRA-VTLA succeeds in 110 of 150 trials (73.33\%), compared with 31.33\%, 50\%, and 44\% for Base, Direct, and adapted FTP-$\pi_{0.5}$ (Table~\ref{tab:simulation}). With matched tactile inputs, success rises from 50\% for Direct to 73.33\% for CoRA-VTLA (95\% paired bootstrap CI for the difference: 14.67 to 32 percentage points, $p_{\mathrm{adj}}=5.1\times10^{-6}$). Paired success is tested with two-sided exact McNemar tests and Holm correction within each benchmark. The largest gain over Base is on plug insertion (0\% to 76\%). Adapted FTP-$\pi_{0.5}$ leads on knob operation (72\%).

\noindent \textbf{Failure outcomes.} CoRA-VTLA reduces timeouts from 96 for Base to 36, while recording four safety failures compared with seven for Base. All four safety failures for CoRA-VTLA occur on nail insertion; plug insertion and knob operation have no safety terminations. These counts are retained in the evaluation denominator.

\noindent \textbf{Generalization to a public benchmark.} To test whether action--response memory generalizes beyond MuJoCo and numeric pressure arrays, the design is adapted to Insert Tube in the public tactile benchmark UniVTAC, which uses a different physics engine, robot, and action interface, and provides optical tactile images instead of pressure arrays or wrist force/torque~\cite{univtac}. A frozen ResNet-18~\cite{resnet} encodes the two fingertip images. Changes in the optical tactile features form a 0.8-second response history, and the action expert retains the current-contact and memory tokens. Base, Direct, and CoRA-VTLA are trained on the same 50 demonstrations and evaluated on the same 100 fixed scenes. CoRA-VTLA reaches 67\% success (Table~\ref{tab:simulation}, last column), compared with 46\% for Base and 51\% for Direct (95\% paired CI for the difference over Direct: 4 to 28 percentage points, $p_{\mathrm{adj}}=0.0195$). Comparing the same scenes, CoRA-VTLA succeeds on 29 trials where Direct fails, while Direct succeeds on 13 where CoRA-VTLA fails. Against Base, the corresponding counts are 36 and 15. The number of slip terminations also falls from 53 with Base and 46 with Direct to 30 with CoRA-VTLA.

\begin{table}[!htbp]
\centering
\small
\setlength{\tabcolsep}{3pt}
\caption{Real-robot success rates (\%).}
\label{tab:real}
\begin{tabular}{lrrrr}
\toprule
Method & Cup & Stamp & Tube & Avg. \\
\midrule
Base ($\pi_{0.5}$) & 38 & 30 & 30 & 32.7 \\
Direct & 50 & 48 & 34 & 44.0 \\
FTP-$\pi_{0.5}$ (adapted) & 46 & 44 & 44 & 44.7 \\
CoRA-VTLA & \textbf{68} & \textbf{76} & \textbf{56} & \textbf{66.7} \\
\bottomrule
\end{tabular}
\end{table}
\noindent \textbf{Real-robot results.} CoRA-VTLA achieves the highest success rate on all three real-robot tasks, averaging 66.7\% (95\% CI [59.3\%, 74.0\%]), compared with 44.7\% for the strongest baseline (Table~\ref{tab:real}). In cup stacking, the grasped paper cup blocks much of the wrist camera's view. The vision-only policy often deforms or drops the cup and sometimes continues pressing after knocking the lower cup out of alignment. In stamping, unstable grasps can cause the stamp to drop during transport, and variations in grasp height change when the stamp touches the paper. A fixed downward stroke can leave an incomplete imprint or continue after contact and trigger a force-limit stop. In tube insertion, the vision-only policy sometimes retracts before reaching the rack and sometimes keeps pushing after the tube jams. Figure~\ref{fig:real-platform} shows CoRA-VTLA withdrawing after a blockage, realigning the tube, and completing the insertion. These episodes illustrate how changes in load after contact can inform whether to continue or adjust the motion.

\begin{figure}[!htbp]
    \centering
    \includegraphics[width=\columnwidth]{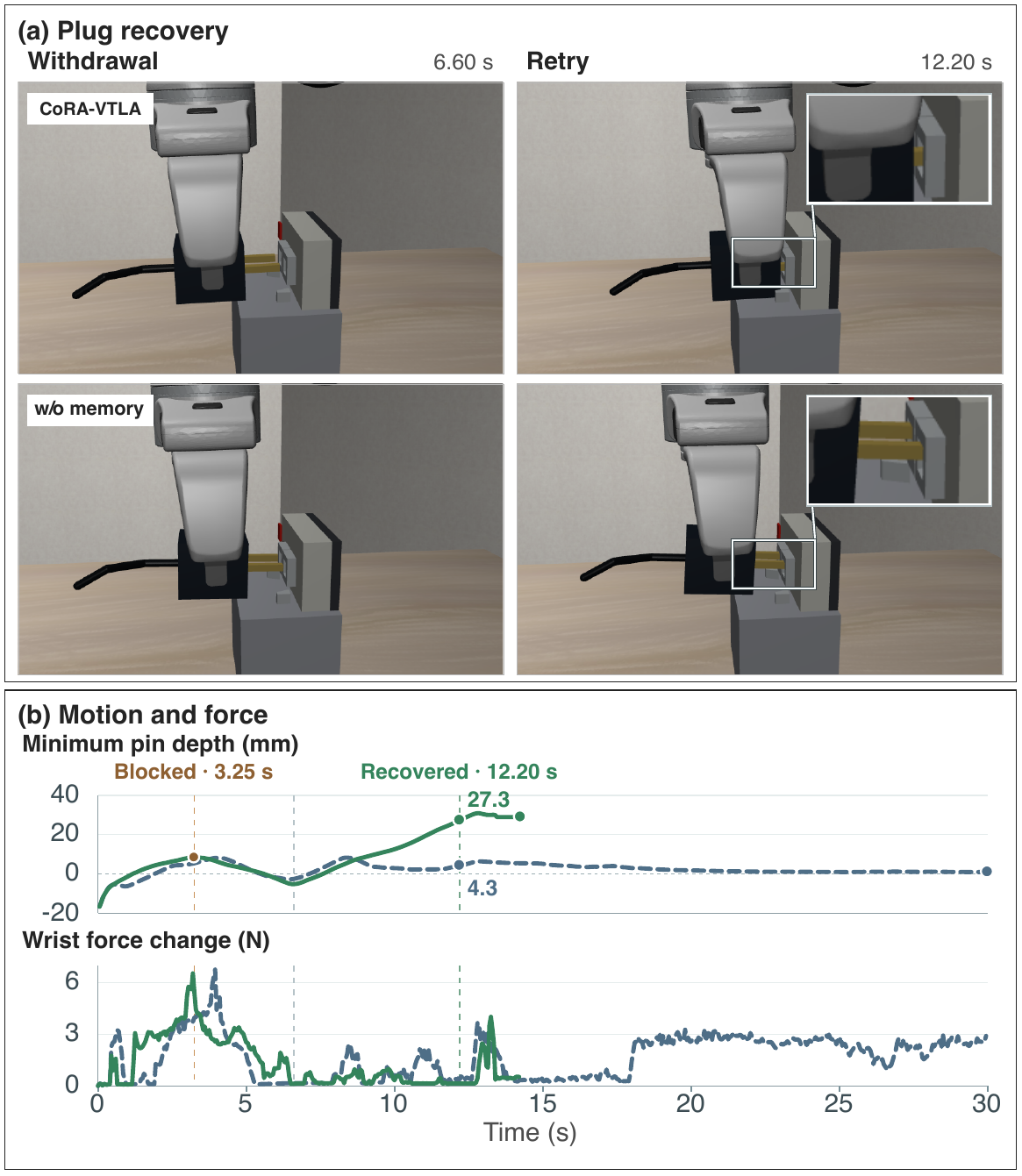}
    \caption{Plug recovery with and without the memory token in a matched MuJoCo scene. The frames show withdrawal and the second insertion attempt, with matched close-ups of the plug and socket. Curves show minimum pin depth and the magnitude of the change in wrist force from the start of each rollout.}
    \label{fig:plug-recovery}
\end{figure}

\begin{figure*}[!t]
    \centering
    \includegraphics[width=\textwidth]{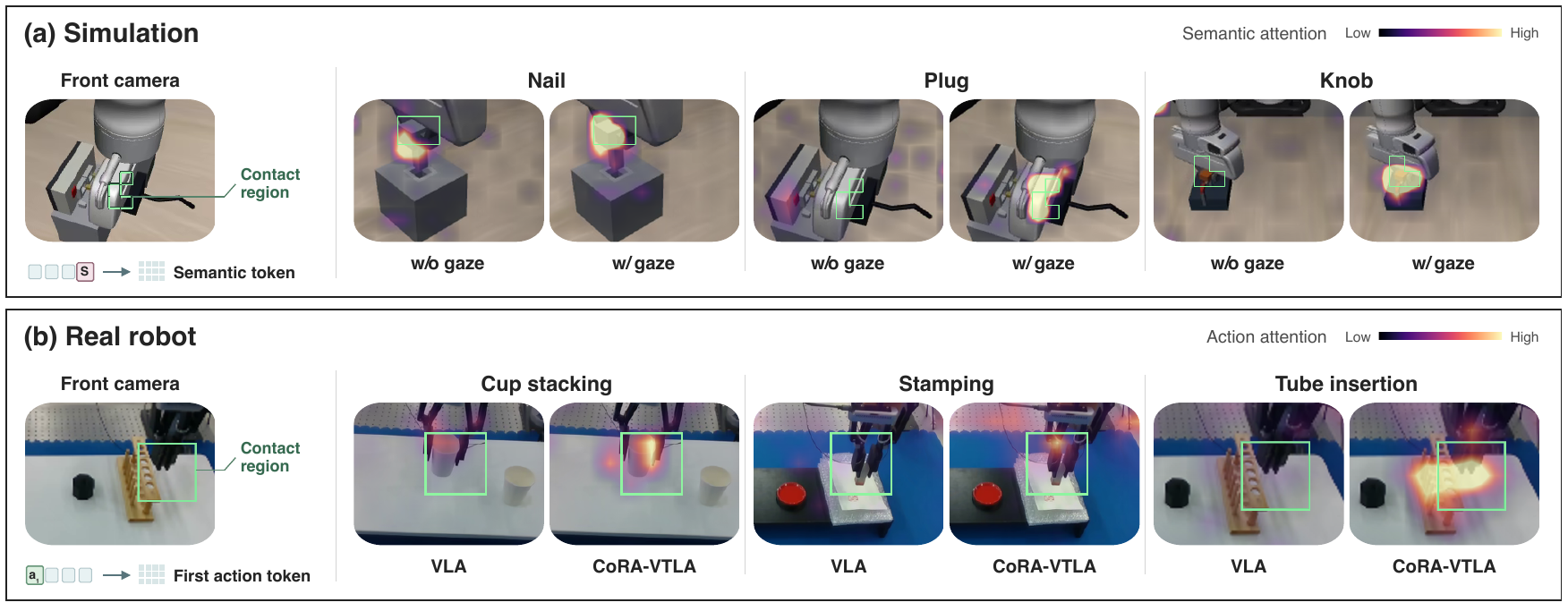}
    \caption{Contact-region attention in simulation and on the real robot. \textbf{(a)} L17 VLM semantic-query attention with and without gaze supervision under the same history configuration on matched MuJoCo frames. \textbf{(b)} First action-token attention at L17 in the final denoising step for VLA and CoRA-VTLA on matched frames from cup stacking, stamping, and tube insertion. Green outlines mark the projected contact regions.}
    \label{fig:contact-attention}
\end{figure*}

\begin{table*}[!t]
\centering
\caption{Ablation study on the MuJoCo tasks (\%). Bold highlights the full model.}
\label{tab:ablations}
\small
\setlength{\tabcolsep}{4pt}
\begin{tabular}{lccccccrrrr}
\toprule
Variant & Command & Motion & Memory & Pretrain & Gaze & Semantic & Nail & Plug & Knob & Avg. \\
\midrule
CoRA-VTLA & -- & -- & \checkmark & \checkmark & \checkmark & \checkmark & \textbf{78} & \textbf{76} & \textbf{66} & \textbf{73.33} \\
CoRA-VTLA + Command \& Motion & \checkmark & \checkmark & \checkmark & \checkmark & \checkmark & \checkmark & 76 & 68 & 60 & 68.00 \\
\midrule
w/o Memory Token & -- & -- & $\times$ & \checkmark & \checkmark & \checkmark & 76 & 48 & 38 & 54.00 \\
w/o Response Pretraining & -- & -- & \checkmark & $\times$ & \checkmark & \checkmark & 76 & 34 & 52 & 54.00 \\
w/o Gaze & -- & -- & \checkmark & \checkmark & $\times$ & \checkmark & 26 & 70 & 52 & 49.33 \\
w/o Semantic Branch & -- & -- & \checkmark & \checkmark & $\times$ & $\times$ & 62 & 54 & 32 & 49.33 \\
\bottomrule
\end{tabular}
\end{table*}

\subsection{Ablation Studies}

Table~\ref{tab:ablations} compares the two history inputs and reports component ablations on the same 150 unseen scenes.

\noindent \textbf{Memory input design.} An initial memory variant included executed control commands and measured robot motion to link attempted actions to actual motion and contact responses. For example, if the robot commands a downward insertion but the end effector barely advances as the contact load rises, these signals together can help the policy detect a blockage. The ablation tests these additional inputs by comparing a memory of commands, motion, and load changes with one that stores only changes in fingertip pressure and wrist force and torque. Their average success rates are 68\% and 73.33\%, respectively, with the largest gap on plug insertion. In a next-step response prediction probe, the MSE with true and shuffled commands is nearly identical (0.0210 for both after rounding). The final model therefore uses the history of load changes. The variant without the memory token achieves 54\% mean success ($p_{\mathrm{adj}}=0.0184$).

\noindent \textbf{Response initialization.} With randomly initialized action-side tactile modules, mean success is 54\% ($p_{\mathrm{adj}}=0.0194$), supporting initialization through response prediction.

\noindent \textbf{Gaze and tactile semantics.} Without gaze supervision, mean success is 49.33\% ($p_{\mathrm{adj}}=0.0020$), with 26\% success on nail insertion. Removing the full semantic branch also yields 49.33\% ($p_{\mathrm{adj}}=0.00015$). With gaze absent in both variants, retaining the semantic branch improves plug insertion (54\% to 70\%) and knob operation (32\% to 52\%), while nail insertion drops from 62\% to 26\%.

\noindent \textbf{Protocol.} The variants share the update budget and evaluation scenes. Paired tests for the ablation comparisons use Holm correction.

\subsection{Qualitative Analysis}
\label{sec:qualitative}

\noindent \textbf{Contact responses and recovery.} Figure~\ref{fig:plug-recovery}(a) shows withdrawal and a second insertion attempt in the same offset scene. Both policies withdraw after the initial blockage. The wrist-force traces in Fig.~\ref{fig:plug-recovery}(b) rise around the blockage and fall during withdrawal. The policies diverge on the second attempt. At 12.20~s, the shallower pin reaches a depth of 27.3~mm with memory and 4.3~mm without it. CoRA-VTLA completes insertion and release at 14.25~s, and the variant without memory times out at 30~s. Across the 25 offset scenes, retaining the memory token increases successes from 8 to 16. Across the 25 aligned scenes, successes increase from 16 to 22. The success gap is larger in offset scenes that require realignment.

\noindent \textbf{Attention under direct tactile fusion.} On 24 matched MuJoCo observations, the normalized spatial entropy of the first action token's main-camera attention at layer 17 rises from 0.838 (Base) to 0.884 (Direct, $p=0.0054$, episode-level sign-flip test). The contact region's share of that attention drops from 1.62\% to 1.18\%. Spatial entropy measures how broadly attention is distributed over the image, while the contact-region share measures how much of that attention reaches the projected contact. Together, the two changes show that Direct spreads attention more broadly while allocating less of it to contact. Figure~\ref{fig:contact-attention}(a) then tests whether spatial supervision can guide the tactile semantic token to the contact region.

\noindent \textbf{Contact-region attention.} With identical history inputs across the two variants, gaze supervision increases the semantic token's mean attention share within the projected contact region at VLM layer 17 from 2.77\% to 46.65\% (Fig.~\ref{fig:contact-attention}(a)). The share increases on all 24 frames ($p=0.00049$, episode-level sign-flip test).

Figure~\ref{fig:contact-attention}(b) shows where the real-robot policy attends when predicting actions. In the matched examples of cup stacking, stamping, and tube insertion, the first action token in CoRA-VTLA assigns more attention to the projected fingertip contact region than the vision-only baseline. In the tube insertion example, the share of front-camera attention within this region rises from 2.6\% to 12.6\% on the same frame. The comparison covers eight tube insertion frames from eight distinct episodes. Both policies receive the same camera observation in each pair. Attention from the first action token is read at layer 17 of the final denoising step, normalized across the front-camera patches, and summed within the same projected contact region for both policies. CoRA-VTLA assigns more attention to the contact region on all eight frames, with the mean share rising from 2.1\% to 7.5\%. The simulation comparison in Fig.~\ref{fig:contact-attention}(a) isolates the effect of gaze supervision on the semantic token, while the real-robot comparison measures how the complete policy attends to contact regions during action prediction.
\section{Conclusion}

\label{sec:conclusion}

CoRA-VTLA adapts pretrained VLAs to precision manipulation through a two-level architecture. At the perception level, contact-guided gaze supervision grounds tactile semantics in the visual scene. At the action level, the action expert generates actions from the VLM context, current contact, and recent responses retrieved from the action--response memory. One-step response prediction on task demonstrations initializes the action-side tactile encoders before joint policy training.

CoRA-VTLA achieves 73.33\% mean success across three MuJoCo tasks, compared with 50\% for direct tactile fusion with matched sensor inputs. With optical tactile sensing, the memory-based policy reaches 67\% on UniVTAC Insert Tube, showing that the memory design transfers across simulators and tactile modalities. CoRA-VTLA also achieves 66.7\% mean success across three real-robot tasks. Extending the memory horizon and improving the safety of recovery remain directions for future work.

% Prevent experimental floats from entering the references.
\FloatBarrier

{
\small
\interlinepenalty=10000
\bibliographystyle{ieeenat_fullname}
\bibliography{main}
}

\end{document}